\documentclass{article}

\usepackage{arxiv}

\usepackage[utf8]{inputenc} % allow utf-8 input
\usepackage[T1]{fontenc}    % use 8-bit T1 fonts
\usepackage{hyperref}       % hyperlinks
\usepackage{url}            % simple URL typesetting
\usepackage{booktabs}       % professional-quality tables
\usepackage{amsfonts}       % blackboard math symbols
\usepackage{nicefrac}       % compact symbols for 1/2, etc.
\usepackage{microtype}      % microtypography
\usepackage{graphicx}

\usepackage{float}
\usepackage{amsmath}
\usepackage{nicealgo}
\usepackage{multirow}
\usepackage{color}
\usepackage{booktabs}

\date{} 					% Or removing it

\usepackage{amsfonts}
\usepackage{amssymb}
\usepackage{amsmath}
\usepackage{algorithm,algpseudocode}
\usepackage{}
\usepackage{url}
\usepackage{graphicx}
\usepackage{color}
\usepackage[dvipsnames]{xcolor}
\usepackage{colortbl}
\usepackage{subcaption}
\usepackage{lineno}
\usepackage{acro}
\usepackage{outlines}

\newcommand{\sh}{20mm}
\newcommand{\sz}{30mm}

\begin{document}

\title{Detecting Unsigned Physical Road Incidents from Driver-view Images}
\author{Alex~Levering \\
	Wageningen University\\
	\texttt{alex.levering@wur.nl} \\
	%% examples of more authors
	\And
	Martin Tomko \\
	University of Melbourne \\
	\And
	Devis Tuia \\
	Wageningen University \\
	\And
	Kourosh Khoshelham \\
	University of Melbourne
	%% \AND
	%% Coauthor \\
	%% Affiliation \\
	%% Address \\
	%% \texttt{email} \\
	%% \And
	%% Coauthor \\
	%% Affiliation \\
	%% Address \\
	%% \texttt{email} \\
	%% \And
	%% Coauthor \\
	%% Affiliation \\
	%% Address \\
	%% \texttt{email} \\
}

% \thanks{Digital Object Identifier: xxx}
% \thanks{TODO: thanks 1}%
% \thanks{TODO: thanks 2}%
% \thanks{Manuscript received TODO:DATE.}

\maketitle

%%%%%%%%%%%%%%%%%%%%%%%%%%%%%% ABSTRACT %%%%%%%%%%%%%%%%%%%%%%%%%%%%%%
\begin{abstract}
\textbf{Note: this is a pre-print version of work accepted in IEEE Transactions on Intelligent Vehicles (T-IV;in press). The paper is currently in production, and the DOI link will be added soon.} \newline
 Safety on roads is of uttermost importance, especially in the context of autonomous vehicles. A critical need is to detect and communicate disruptive incidents early and effectively. In this paper we propose a system based on an off-the-shelf deep neural network architecture that is able to detect and recognize types of unsigned (non-placarded, such as traffic signs),  physical (visible in images) road incidents. We develop a taxonomy for unsigned physical incidents to provide a means of organizing and grouping related incidents. After selecting eight target types of incidents, we collect a dataset of twelve thousand images gathered from publicly-available web sources. We subsequently fine-tune a convolutional neural network to recognize the eight types of road incidents. The proposed model is able to recognize incidents with a high level of accuracy (higher than 90\%). We further show that while our system generalizes well across spatial context by training a classifier on geostratified data in the United Kingdom (with an accuracy of over 90\%), the translation to visually less similar environments requires spatially distributed data collection.
\end{abstract}

%%%%%%%%%%%%%%%%%%%%%%%%%%%%%%%%%%%%%%%%%%%%%%%%%%%%%%%%%%%%%%%%%%%%%%%%%%
%%%%%%%%%%%%%%%%%%%%%%%%%%%%%% INTRODUCTION %%%%%%%%%%%%%%%%%%%%%%%%%%%%%%
\section{Introduction}

Roads are a highly dynamic environment, continuously impacted by ephemeral changes to surface conditions. Incidents that interrupt the road network reduce network connectivity cause economical damage by increasing travel time, causing delays in deliveries, and lead to missed connections to other modes of transport. In England, congestion caused by incidents on trunk roads and motorways is estimated to cost approximately half a billion pounds every year~\cite{highways_england_smart_2016} and delaying approximately one in five journeys~\cite{peluffo_strategic_2015}. This will worsen, as traffic on England's trunk roads and motorways has grown by over 50\% since 1993 and is expected to grow another 31\% by 2041~ \cite{highways_england_highways_2017}. Combined with the continued trend of car ownership worldwide~\cite{dargay_vehicle_2007} and a continued increase in road vehicle miles~\cite{peluffo_strategic_2015}, the effects of incidents on the serviceability of the road network are likely to deteriorate. The low-cost and wide availability of vehicle-borne cameras offer great potential to visually detect road incidents. This information could subsequently be communicated across the networked traffic. Yet, previous research on incident perception and detection from the vehicle-centric perspective (ego-vehicle) has thus far not widely considered the detection of incidents. In this research we propose a taxonomy of applicable incidents, collect a first-of-its-kind machine-learning dataset of incident images approximating a vehicle-centric perspective, and develop and test a deep learning model to recognize unsigned physical incidents.

The rest of this paper is organized as follows:
Section~\ref{sec:Background} gives an overview of the background material of the research. Section~\ref{sec:UnsigedPhysical} discusses the definition of unsigned physical incidents and the taxonomy of incidents. Section~\ref{sec:Methodology} presents the methodology and experimental set-up. In Section~\ref{sec:Results} we discuss the results of the model classification, and discuss its nuances in Section~\ref{sec:Discussion}. We draw conclusions from our work in Section~\ref{sec:Conclusions}.

\section{Background}\label{sec:Background}
\subsection{Definitions of incidents}
To properly study unsigned physical incidents, the concept must first be defined. The United States Federal Highway Administration defines an incident as \textit{"any non-recurring event that causes a reduction of roadway capacity  or  an  abnormal  increase  in  demand."}~\cite[p.2]{p.b._farradyne_traffic_2000}. This definition is unsuitable as it does not account for a reduction in serviceability. Secondly, it attempts to limit incidents to non-recurring events, which excludes events such as snowfall. Berdica defines an incident as \textit{"an event, which directly or indirectly can result in considerable reductions or interruptions in the serviceability of a link/route/road network."}~\cite[p.118]{berdica_introduction_2002}. This definition is well-suited for this research, as it covers the reduction in serviceability and does not limit the context in which incidents may occur. We thus apply Berdica's definition as a basis for defining incidents in this paper. We further delineate incidents by their physical nature. We consider physical incidents to be incidents that are perceivable by sensors such as cameras, sonar systems, and laser scanners. Digital events such as a traffic light hack are not considered, because the cause of the disruption is not easily interpretable by in-situ sensors.

\begin{figure}[ht]
  \begin{center}
    \includegraphics[width=0.5\textwidth]{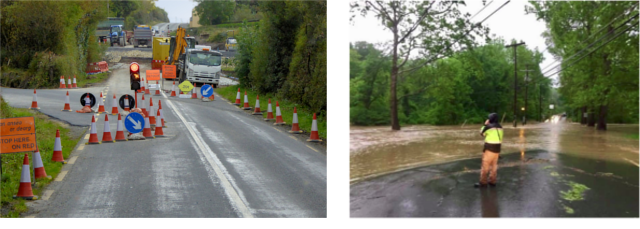}
    \caption{Left: A signed physical incident with signposted roadworks \cite{dixon_roadworks_2017}. Right: An unsigned physical incident -- road flooding \cite{sandy_spring_volunteer_fire_department_ssvfd_2017}.}
    \label{fig:inc_types}
  \end{center}
\end{figure}

Finally, we further distinguish between \textit{signed} and \textit{unsigned} incidents. Signed incidents are incidents which are signposted or otherwise marked as a hazard. For instance, roadworks and parades are often signposted with barriers, traffic signs, and high-visibility equipment. On the contrary, unsigned incidents are unplanned, such as a flash-flood after heavy rainfall. Figure~\ref{fig:inc_types} gives an example of both types of physical incidents. Once detected and acted upon by authorities, incidents are often signposted. In this research we only consider unsigned physical incidents. We do not consider signed physical incidents because it shares a broad overlap with tasks such as street sign recognition \cite{zhu_traffic-sign_2016, zaklouta_real-time_2014}, traffic cone recognition \cite{yong_real-time_2015}, road marking detection \cite{hillel_recent_2014}, and roadworks detection \cite{fazekas_locating_2017}.

\subsection{Image-based Incident Detection}
Animals are a frequent cause of collisions with cars. Research by Zhou et al~\cite{zhou_real-time_2014} considers the state of animal detection systems for road-going vehicles until 2014. They record five papers applicable to intelligent vehicles' need to classify various species of animals. Most of these efforts concern general animal classifiers rather than classifiers specialized to the detection of animals on road. Directly relevant to unsigned physical incident detection is the semantic segmentation in synthetic images of kangaroos on the road for collision prevention~\cite{saleh_kangaroo_2016}. This work relied on training on computer-generated images of kangaroos. Related work on road damage detection from smartphone dashcam images \cite{alfarrarjeh_deep_2018} did focus on road surface assessment rather than damage that may cause road closure.

Other types of unsigned physical incidents are scarcely considered in existing literature. Chen et al~\cite{chen_lidar-histogram_2017} consider the use of LIDAR sensors for the detection of road obstacles by defining the driving surface and then detecting outliers the surface. In~\cite{levi_stixelnet:_2015}, this same task is approached through images from true color cameras to detect perceivable edges of the driving surface and thus find the ground plane boundaries of potential obstacles. However, neither of these works attempts to classify the object that is obstructing the driving path, nor do they distinguish regular obstacles (e.g., cars) from unusual obstacles (e.g., debris). To our best knowledge, the only research in this domain that defines the specific type of hazard to occur on the driving surface is \cite{shao_research_2015}, in which authors detect shallow holes and water hazards on the driving surface using the (lack of) returns of a given LIDAR beam. Similar research based on RGB images is still lacking.

\subsection{Computer Vision Techniques}
Many applicable feature extraction techniques exist, such as first \& second-order edge detection, image motion descriptions, shape matching, texture extraction, and statistical features~\cite{s._nixon_feature_2012}. Convolutional Neural Networks (CNNs) were conceptualized by Fukishima \cite{fukushima_neocognitron:_1980} and expanded upon by LeCun et al \cite{lecun_gradient-based_1998} as a means to automatically learn features from images by using trainable filters. Krizhevsky et al~\cite{krizhevsky_imagenet_2012} implemented a GPU-based CNN which laid the foundation for modern deep CNN network architectures. VGG~\cite{simonyan_deep_2014} improved upon the AlexNet architecture by experimenting with layer configurations and explored deeper networks. ResNet \cite{he_deep_2016} made notable improvements to CNNs by introducing \textit{skip connections}. Skip connections let network layers learn identity mappings enabling to train very deep networks while accounting for the vanishing gradient problem \cite{hochreiter_vanishing_1998}. Additional batch normalization layers enabled to stabilize the signal of intermediate layers. Further improvements to CNN architectures are continuously made, but for brevity are not discussed here.

\section{UNSIGNED PHYSICAL INCIDENTS}\label{sec:UnsigedPhysical}
By enhancing the definition of incidents proposed by Berdica with the concepts of physicality and signage, the working definition for unsigned physical incidents used in this research is:
\textit{An unsigned physical incident is an event without road signs or hazard markers that can be detected by sensors, which directly or indirectly can result in considerable reductions or interruptions in the serviceability of a link/route/road network.}

\begin{figure*}
    \centering
    \includegraphics[width=0.5\textwidth]{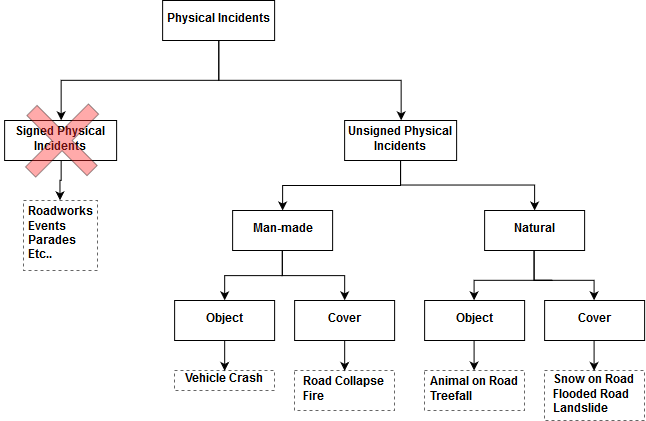}
    \caption{Taxonomy of incidents and their Semantic groupings}
    \label{fig:taxonomy}
\end{figure*}

\subsection{Taxonomy of Incidents}
There are many incidents which may affect a road network, many of them related through shared characteristics. For instance, snowy and inundated roads are distinct phenomena, but related as they both concern natural events affecting the road surface. In order to systematically explore the breadth of the domain of possible incident types before developing a classifier, we first propose a taxonomy that semantically groups incidents. We define the structure of the taxonomy through Formal Concept Analysis (FCA) \cite{ganter_formal_2012} in order to uncover attributes and attribute groupings of incidents. We iteratively refined and grouped attributes until they provided hierarchical, binary semantic groupings. From the FCA we identified two common aspects distinguishing between incidents. The first aspect is the manifest and most likely cause of an incident, i.e., man-made or natural causes. For instance, a car crash involving two cars manifests as man-made obstructions on the road. Whether it is caused by a driver error or a mechanical failure, both causes are rooted in human failure. Flooding on the other hand manifests as a natural cause, although it may be triggered by a human damaging a road hydrant. This aspect must be interpreted as the perceivable nature of the incident. The second grouping is whether the incident is by nature a well-defined discrete (set of) objects or a continuous field, i.e., a \textit{cover}. Road flooding is a continuous phenomenon without a well-defined discrete delineation of the incident. On the other hand, a fallen tree can be counted and can be considered a discrete incident, which we refer to as \textit{objects}. We show how the resulting taxonomy may be structured formally in Figure \ref{fig:taxonomy}. Such a distinction should serve to inform the comprehensive, systematic collection of training data. It may further function as a test on the model's capacity to classify groups of incidents. As more refined models are produced, the taxonomy may deepen to include more probable causes to provide a more fine-grained risk estimation.

The taxonomy has been been designed with several limitations in mind:
\begin{itemize}
  \item While an error in a traffic management system may cause traffic disruption, it is hard if not impossible to detect this incident by sensors in-situ. Hence, digital incidents are not included. In a full hierarchy of possible incidents it would occur on the same level as physical incidents;
  \item We consider attributes of incidents by their perceivable cause. It is possible that e.g., a tree was purposely cut to block a road, making it a man-made adversarial incident. Yet, such an incident is often hard to distinguish from natural tree fall, a much more common occurrence;
  \item Signed physical incidents are not considered in this research as the detection of signage is actively researched using computer vision techniques. With sufficient research progress the detection of road signs may soon cover for the detection signed incidents and events such as roadworks and parades.
\end{itemize}

\subsection{Incident Definitions}
The lowest-level labels refer to individual incidents that belong to the specified groupings. Here, we consider eight class labels. We chose a variety of incidents that can be semantically described for the purposes of collecting a training dataset of images and for which we can reasonably expect to find hundreds of images per class, covering the spectrum and variability of common objects and surfaces wherever possible. We use the following working definitions of classes in this research:

\begin{itemize}
\item \textbf{Animal on Road}: Any animal, both living and dead, situated on or within close proximity of the driving surface
\item \textbf{Collapse}: A major break-up of the driving surface which would be too big for common motor vehicles to drive across without incurring damage
\item \textbf{Fire}: An uncontrolled and active fire anywhere in the image that may affect the driving conditions immediately or when left uncontrolled
\item \textbf{Flooding on Road}: A (section of) driving surface that is submerged in a cover of water puddles such that it causes drivers to change their driving behaviour
\item \textbf{Landslide}: A cover of dirt, rocks, or natural debris originating from a raised surface, which has settled on the driving surface
\item \textbf{Snow on Road}: Any amount of snow on the driving surface such that it could cause drivers to change their driving behaviour
\item \textbf{Treefall}: A tree, trunk, or sizable branch leaning over or lying on the driving surface in such a way that it would obstruct traffic
\item \textbf{Vehicle Crash}: Any visible collision between one or more motor vehicles, or a motor vehicle collision with an object in the environment, such as a tree
\end{itemize}

\begin{figure*}[!ht]
\centering
\begin{tabular}{cccc}
 \includegraphics[width=\sz,height=\sh]{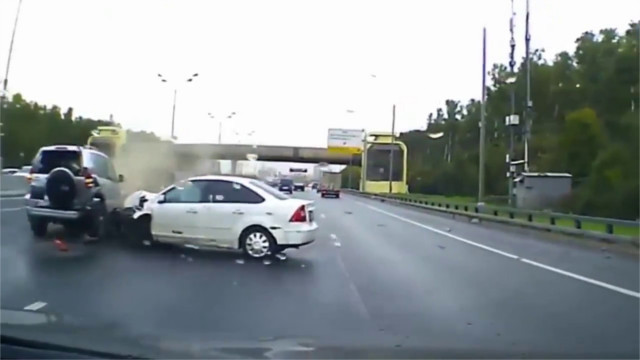}&
\includegraphics[width=\sz,height=\sh]{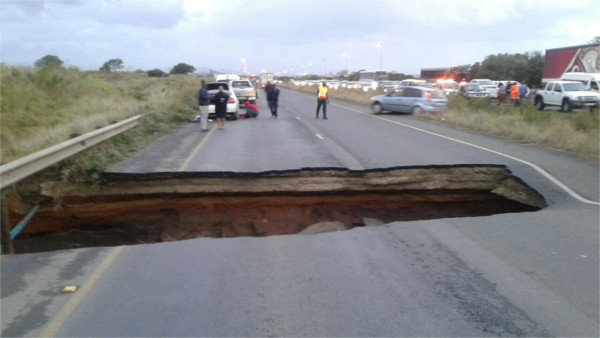}&
\includegraphics[width=\sz,height=\sh]{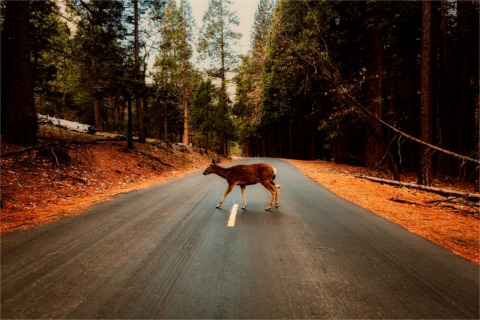}&
\includegraphics[width=\sz,height=\sh]{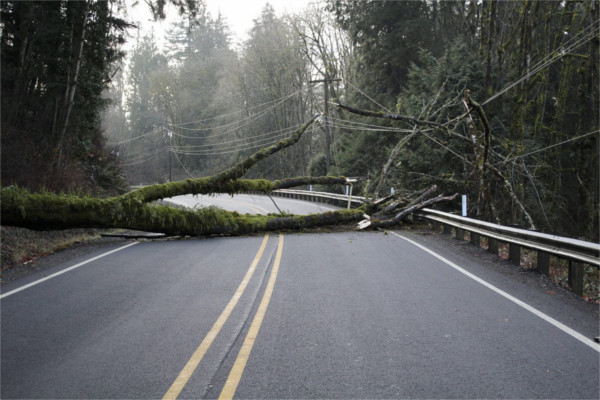}\\
(a) Crash & (b) Collapse & (c) Animal & (d) Treefall \\
 \includegraphics[width=\sz,height=\sh]{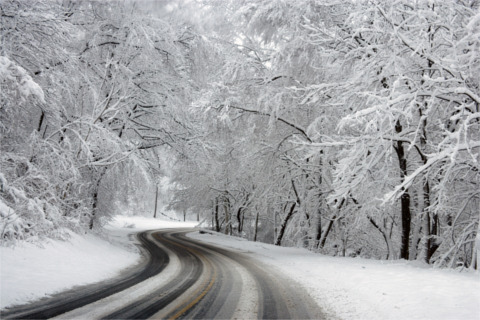}&
\includegraphics[width=\sz,height=\sh]{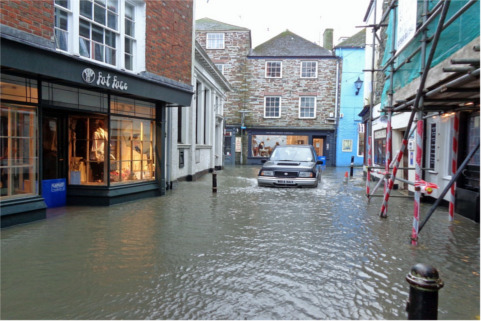}&
\includegraphics[width=\sz,height=\sh]{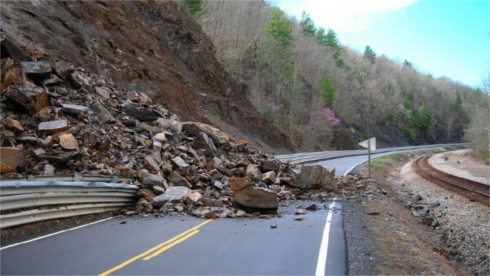}&
 \includegraphics[width=\sz,height=\sh]{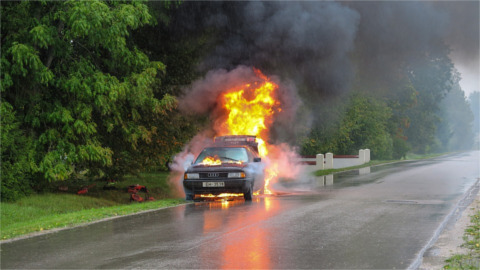}\\
(e) Snow& (f) Flood & (g) Landslide & (h) Fire\\
 \end{tabular}
\caption{Prototypical images of each incident class covered in this research}
\label{fig:prototypes}
\end{figure*}

\newpage
\section{METHODOLOGY}\label{sec:Methodology}
We perform supervised single-label classification on web-gathered positive examples and a diverse set of negative examples sampled from various sources. We do not consider multilabel-classification to be fit for the case, as multiple incidents occurring at once are unusual.

\subsection{Image Dataset}\label{sec:constr}
The input dataset of imagery of various incidents occurring on the road network is built from a multitude of sources. Each image is attributed to only one of the fine-level classes via a single label. Thus, images that contain, e.g., both a crash and a fire are excluded from the training set. Images were selected to accurately represent road scenes from the perspective of a vehicle. Images should be taken from the vehicle-perspective with expected viewports. For instance, an image taken just centimeters above the road is not representative for a driving scene. Likewise, an image that is rotated beyond a few degrees is not relevant either. Since this information is not supplied with most images it is up to human labellers to filter scenes to match these criteria.

\subsubsection{Positive Examples}
In order for a classifier to differentiate between incidents and normal driving situations, a dataset of positive examples has to be gathered. Positive examples of unsigned physical incident images have one of the incidents clearly in view. We gather the set of positives from four sources. Firstly, we perform Web harvesting from Google \cite{google_custom_nodate}, Flickr \cite{yahoo_flickr_nodate}, and Bing \cite{microsoft_bing_nodate} image APIs. Per query, we retain the first 100 images as the quality of returned images quickly declined after this limit. We gather images by constructing queries using synonym pairs. For two sets of predefined keywords, we combine each set pairwise to form a new set with pairs two keywords each. For instance, the sets \{\textit{road, street}\} and \{\textit{snow, blizzard}\} are combined to form the set \{\textit{road snow, road blizzard, street snow, street blizzard}\}. We also constructed additional queries based on the translation of these English keyword sets into Dutch, Croatian, Farsi, Mandarin, and Slovak. We do so to increase the diversity of scenes and to reduce the influence of geographic bias. Lastly, we filter duplicate returns by checking for absolute equivalence between images matrices. Prototypical examples of each incident type are given in Figure~\ref{fig:prototypes}

\begin{figure*}[!t]
  \centering
    \includegraphics[scale=1.75]{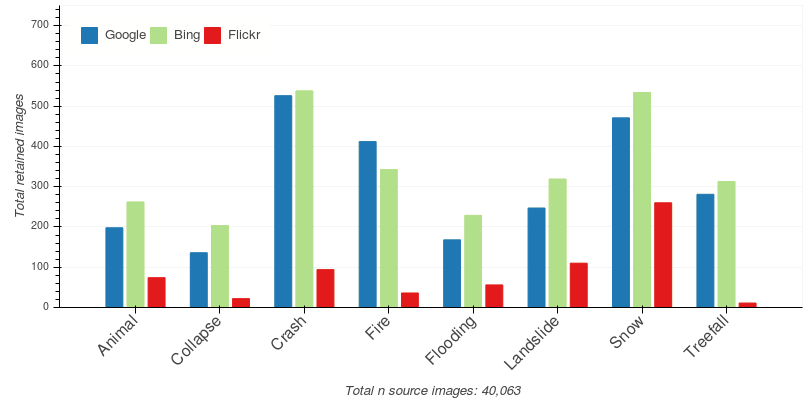}
    \caption{Overview of images per class as derived from each source, harvested using English language queries.}
    \label{fig:harvested}
\end{figure*}

\begin{figure*}[!t]
  \centering
    \includegraphics[scale=1.75]{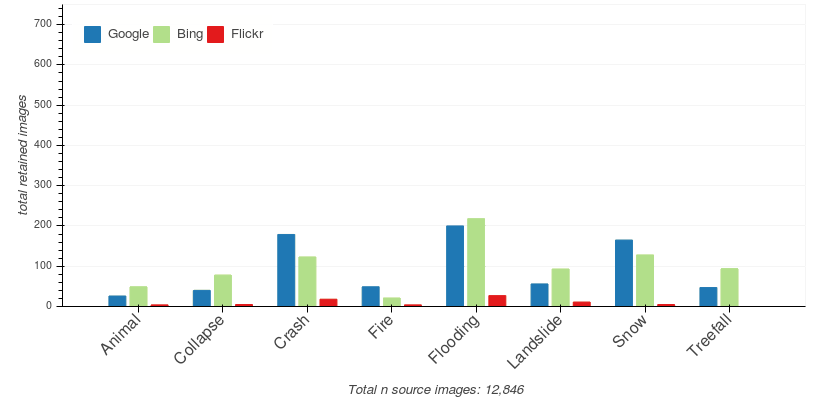}
    \caption{Overview of images per class as derived from each source using the non-English queries.}
    \label{fig:multilingual}
\end{figure*}

Through synonym combinations, we formed a set of 118 English language queries which retrieved 40,063 images. After selecting suitable examples, 5,844 images were included in the dataset. We retain 2,439 images from Google, 2,742 from Bing, and 663 from Flickr. Note that duplication filtering works in favour of the total amount of images retained from Bing. The total amount of duplicates removed was not tracked. Figure~\ref{fig:harvested} lists the distribution of images retained per class along with their sources. Google and Bing provide a superior rate of correct images when compared to Flickr, while being highly similar in the amount of correctly returned images per class. By translating the 118 representative queries into the five non-English languages, we performed additional 63 queries which retrieved 12,846 images, of which 1,641 were included in the dataset. We retained 762 images from Google, 804 from Bing, and 74 from Flickr. Figure \ref{fig:multilingual} displays the distribution of images retained after non-English queries, per class and by source.

To add to the web-sampled images, the Geograph UK project supplied a database export for three classes, namely \textit{animal on road}, \textit{flooding}, and \textit{snow}. The database exports were made using the search terms \textit{animals/cow/sheep on road}, \textit{flooding road}, and \textit{snow road}. Table \ref{tab:gathertype} provides an overview of positive collected images as gathered from English and non-English Web harvesting queries and from Geograph.

\begin{table}[bt]
\caption{Total amount of images per class as collected by gathering type.}
\label{tab:gathertype}
\centering
\begin{tabular}{l | rrr | r}
\toprule
\textbf{Incident} & \textbf{English} & \textbf{non-English} & \textbf{Geograph} & \textbf{Total} \\
\hline
Animal on road & 534   	& 79  	& 708 	& 1,321  \\
Collapse       & 362   	& 123 	& 6   	& 491    \\
Crash          & 1,158 	& 320 	& -   	& 1,478  \\
Fire           & 791   	& 74	& -		& 865    \\
Flooding       & 453	& 446   & 1,257 & 2,156  \\
Landslide      & 676	& 149	& -		& 825    \\
Snow           & 1,265	& 304	& 3,174	& 4,743  \\
Treefall       & 605	& 146	& -		& 751    \\
\midrule
\textbf{Total} & 5,844  & 1,641	& 5,145	& 12,630 \\
\bottomrule
\end{tabular}
\end{table}

\subsubsection{Negative Examples}
Since this research is concerned with recognizing incidents as compared to normal driving situations, a dataset of negative examples was sampled. Negative examples are images of driving situations without an incident affecting the road. For this purpose we sample from various sources to cover the preconditions of the research:

\begin{itemize}
\item \textbf{Berkeley Deep Drive (20k):} a random sample of 20k images from the 100k image subset of the Berkeley Deep Drive (BDD) dataset~\cite{seita_bdd100k:_2018}. This dataset consists of dashcam footage captured in a variety of US cities. It is notable for the variety of captured scenes and weather conditions. The inclusion of scenes containing wet and snowy conditions makes this dataset especially relevant to help distinguish between disruptive and non-disruptive weather conditions. It also contains variations in rotation and angle similar to the positive dataset. The image dataset contains the 10th frame of each video in their 100k videos dataset. This dataset thus captures the widest variety of driving conditions of the negatives datasets such as weather conditions, geographic diversity (across the United States), day/nighttime, and complications, e.g., reflections of the dashboard.

\item \textbf{Cityscapes (10k):} sample images from the Cityscapes dataset~\cite{cordts_cityscapes_2016} improve the geographic diversity of our dataset. This dataset covers a variety of street scenes of German cities and contains every 20th frame of 30fps video sequences. While less varied in weather and camera conditions, the driving conditions in the dataset are distinctly more European than the BDD dataset. We include 10k images from the Cityscapes dataset to improve the geographic diversity of the negatives class.

\item \textbf{Geograph (10.2k):}: a random sample of 10.2k images tagged as \textit{'road transport'} from the Geograph project \cite{noauthor_geograph_2018}. The dataset contains unfiltered stills covered by the tag, and therefore reflects a high diversity of images, at times even irrelevant to driving conditions in general. While most photos are taken from a viewpoint similar to BDD and Cityscapes, a number of images also contain odd angles and targets (e.g. streams or pastures). We include this dataset to offset the strong urban focus of the benchmark datasets, as well as to counteract potential overfitting on the difference in viewpoint rotation, angle, and orientation in the incidents dataset. We retain 10k images from this dataset to ensure the inclusion of landscape images, which are frequent in the positives dataset.
\end{itemize}

Images from the BDD dataset often contain elements of the ego vehicle in the image itself (e.g., dashboard or bonnet). To reduce the chance of overfitting the negatives dataset onto such irrelevant visual cues, we crop out the bottom 25\% of all images from the BDD dataset. To retain the image aspect ratio, we also crop 12,5\% from both the left and right sides of the image. Images are resized to 224x224 pixels to match the input size of the classification model. We subset the images into batches of 70/20/10\% of the full data for the training, validation, and testing split respectively.

\subsection{Set-up of CNN Model}
To perform incident detection we fine-tune a ResNet-34 model~\cite{he_deep_2016} pre-trained on the ImageNet dataset~\cite{deng_imagenet:_2009}. To do so, we unfreeze all layers of the network. During training we consider the individual classes listed at the lowest-level nodes in Figure~\ref{fig:taxonomy}. To account for the imbalance in the number of positive and negative samples and to limit overfitting onto specific classes we scale the loss according to the inverse frequency of the number of images of each class across the dataset, as in Equation~\ref{eq:loss}.

\begin{equation}\label{eq:loss}
    Loss = Loss *  \left(1 -\frac{\sum_{i=1}^{c} c_i}{\sum_{i=1}^{C} C_i}\right)
\end{equation}{}

Here, C denotes the set of all image labels (including negatives) in the dataset, and $c \in C$ is the set of positive examples of the ground truth class of the input image. We train the model with the parameter settings listed in table \ref{tab:params}. We optimize the model using the RMSprop optimization algorithm \cite[s.29]{hinton_neural_2015} without applying momentum.

\begin{table}[h]
    \caption{Classification performance of the best model trained on the complete dataset}
    \label{tab:params}
\centering
    \begin{tabular}{ll}
    \toprule
    \textbf{Parameter}           & \textbf{Value}             \\ \midrule
    Batch size                   & 10                         \\ 
    Initial learning rate        & 0.0001                     \\ 
    Learning rate decay schedule & Decay at epochs 10, 30, 40 \\ 
    L2 regularization strength   & 0.0001                     \\ 
    \bottomrule
    \end{tabular}

\end{table}

We apply the following random augmentations with a 50\% chance of occurrence where applicable:

\begin{itemize}
    \item Random horizontal flip
    \item Random greyscale transform
    \item Random rotation up to 5 degrees in either direction
    \item Jittering of hue/brightness/contrast/saturation up to a factor of 0.05
\end{itemize}

Last, we normalize all images to the mean and standard deviation of the colour bands of the training subset. Network training is performed using PyTorch version 0.3 running on Python version 3.6, using the freely-available Google Colaboratory \cite{google_google_2018}.

\section{Results}\label{sec:Results}

\subsection{Full Dataset Results}
\begin{figure*}[!t]
  \centering
    \includegraphics[width=0.6\textwidth]{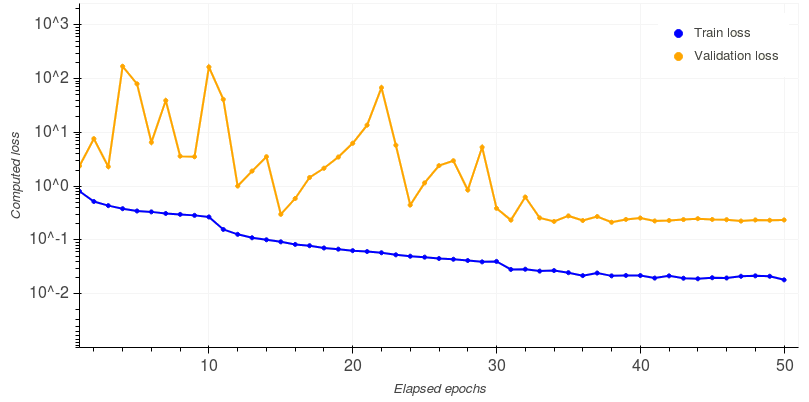}
    \caption{Loss curve of the model trained on the complete dataset.}
    \label{fig:full_loss}
\end{figure*}

Training was concluded after 50 epochs, with a stable state being reached after 37 epochs. Figure~\ref{fig:full_loss} displays the loss for the model at every epoch during training and validation, while Table~\ref{tab:all_acc} shows the final accuracy and average F1-score derived for each phase on the best model. The confusion matrix for the testing phase is given in Table~\ref{tab:all_test}, showing the expected trend that most misclassifications pertain to the flooding class.

\begin{table}[h]
    \caption{Classification performance of the best model trained on the full dataset}
    \label{tab:all_acc}
    \centering
    \begin{tabular}{l | ccc}%{| l | S[table-format=2.4]S[table-format=2.4]S[table-format=2.4] |}
        \toprule
        \textbf{Metric}   					& \textbf{Training} & \textbf{Validation} & \textbf{Testing} \\
        \midrule
        \textbf{Accuracy} 					& 99.49\% & 96.31\% & 97.15\% \\
        \textbf{F1-score} 	& 0.9403 & 0.9054 &  0.8909 \\
        \textbf{Loss}     					& 0.02149 & 0.2135 & 0.1761 \\
        \bottomrule
    \end{tabular}
\end{table}

\begin{table*}[!t]
    \caption{Testing split confusion matrix ($n$=5,263) of the best model trained on the full dataset. The rows represents the true class and the columns represents the predicted class}    
    \label{tab:all_test}
    \centering

    \begin{tabular}{ l | rrrrrrrrr | l | l }
    \toprule
%    \backslashbox{\textbf{True}}{\textbf{Predicted}} &     & \multicolumn{8}{l}{} \vline & \pbox{20cm}{\textbf{F1-} \\ \textbf{Measure}} & \pbox{20cm}{\textbf{Top-1} \\ \textbf{Accuracy}} \\
\textbf{True label} & \multicolumn{9}{c|}{\textbf{Predicted}} & \textbf{F1} & \textbf{Top-1}\\
    \midrule
    \textbf{Animal on Road} & 129  & 0   & 1    & 0  & 2   & 1  & 0    & 1   & 1  	& 0.9021 & 95.56\% \\
    \textbf{Road Collapse}  & 0    & 50  & 1    & 0  & 0   & 0  & 0    & 0   & 3  	& 0.9174 & 92.59\% \\
    \textbf{Vehicle Crash}  & 1    & 0   & 155  & 0  & 0   & 0  & 1    & 1   & 2  	& 0.9394 & 96.88\% \\
    \textbf{Fire}        	& 0    & 0   & 0    & 97 & 0   & 1  & 0    & 0   & 2  	& 0.9848 & 97.00\% \\
    \textbf{Flooded Road}   & 0    & 1   & 0    & 0  & 188 & 0  & 1    & 2   & 20 	& 0.8806 & 88.68\% \\
    \textbf{Landslide}    	& 0    & 1   & 0    & 0  & 0   & 65 & 1    & 0   & 3  	& 0.9028 & 92.86\% \\
    \textbf{Treefall}       & 0    & 0   & 1    & 0  & 1   & 2  & 67   & 1   & 1  	& 0.9241 & 91.78\% \\
    \textbf{Snow on Road}   & 2    & 0   & 0    & 0  & 3   & 0  & 0    & 468 & 14 	& 0.9689 & 96.10\% \\
    \textbf{Negative}       & 19   & 3   & 12   & 0  & 21  & 6  & 2    & 6   & 3894 & 0.9854 & 98.26\% \\
    \bottomrule
    \end{tabular}
 \end{table*}

\subsection{Geographical Stratification}
To assess the influence of geographical correlation in visual features in images on the incident detection task, we explore the impact of geographical stratification of the training image set. We run an experiment using three incident classes present in the Geograph data: \textit{Animals}, \textit{Flooding}, and \textit{Snow}, while considering \textit{negatives} as usual. We only use these three positive classes and the full dataset of negatives as the images retrieved from the Geograph project have reliable geotags that are situated in the United Kingdom or Ireland, and as such they can be regionally stratified by locations in these two countries. Images from England, Scotland, or Ireland are included in the training or validation dataset, while images from Wales form the holdout dataset. We thus effectively split the Geograph data situated in England, Scotland, and Ireland to a 72.5/22.5/5\% split, with the 5\% holdout data situated in Wales so that we can test the trained model performance on unseen data from a new geographical region.

During training and validation of the geographically stratified dataset we include the harvested and non-English data. 75\% of the harvested and non-English data in each of the three positive incident classes is added to the training dataset, while the remaining 25\% data is distributed to the validation dataset. While it is possible that some of the harvested data are situated in Wales, UK, the high ratio of Geograph to non-Geograph data in this class (2:1 for all three classes) assures that the occurrence of undesirable geographical correlation resulting from the inclusion of harvested images is low. Model training adhered to the same parameters as for all incident classes, as the chosen hyperparameters were observed to lead to good convergence.

Training was concluded after 50 epochs, while validation loss stabilized after 33 epochs. Table~\ref{tab:geo_acc} displays the final accuracy and F1-score derived for each phase on the best model. The confusion matrices for the testing phase is given in Table~\ref{tab:geo_test}. The trends visible in the confusion matrix largely follow the trend seen for the complete dataset. While overall the accuracy is good, the drop in accuracy and F1-score is notable.

\begin{table}[h]
    \caption{Classification performance of the best model trained on the geo-stratified dataset.}
    \label{tab:geo_acc}
    \centering
    \begin{tabular}{l|lll}
    \toprule
    \textbf{Metric}   & \textbf{Training} & \textbf{Validation} & \textbf{Testing} \\
    \midrule
    \textbf{Accuracy} & 97.90\% & 96.59\% & 92.90\% \\
    \textbf{F1-score} & 0.9403 & 0.9054 & 0.9169 \\
    \textbf{Loss}     & 0.0771 & 0.1352 & 0.1973 \\
    \bottomrule
    \end{tabular}

\end{table}

\begin{table}[h]
    \caption{Testing split confusion matrix ($n$=309) of the best model trained on the geographically stratified dataset. The rows represents the true class and the columns represents the predicted class.}
    \label{tab:geo_test}
    \centering
    \begin{tabular}{ l | llll | l | l }
    \toprule
%    \backslashbox{\textbf{True}}{\textbf{Predicted}} & 
 %   & \multicolumn{3}{l}{} \vline & \pbox{5cm}{\textbf{F1-} \\ \textbf{Measure}} & \pbox{5cm}{\textbf{Top-1} \\ \textbf{Accuracy}} \\
 \textbf{True label} & \multicolumn{4}{c|}{\textbf{Predicted}} & \textbf{F1} & \textbf{Top-1}\\

    \midrule
    \textbf{Animals on Road} & 73	& 0     & 0    & 0    & 0.9299 & 100\%   \\
    \textbf{Flooded Road}    & 1	& 54    & 3    & 0    & 0.9319 & 93.10\% \\
    \textbf{Negative}		 & 10   & 3     & 48   & 2    & 0.8205 & 76.19\% \\
    \textbf{Snow on Road}    & 0	& 0 	& 3    & 112  & 0.9782 & 97.39\% \\
    \bottomrule
    \end{tabular}

\end{table}

\section{DISCUSSION}\label{sec:Discussion}
\subsection{Incident Recognition}
\begin{figure*}
\centering
\begin{tabular}{cccc}
 \includegraphics[width=\sz,height=\sh]{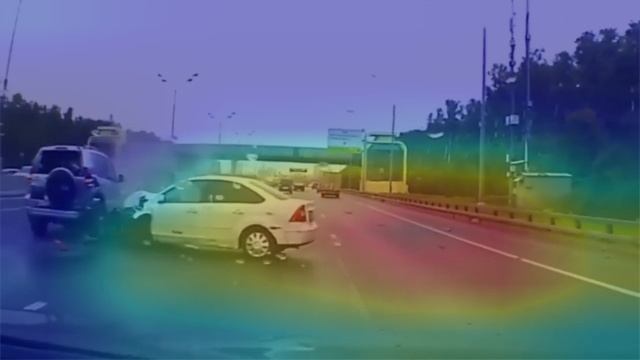}&
\includegraphics[width=\sz,height=\sh]{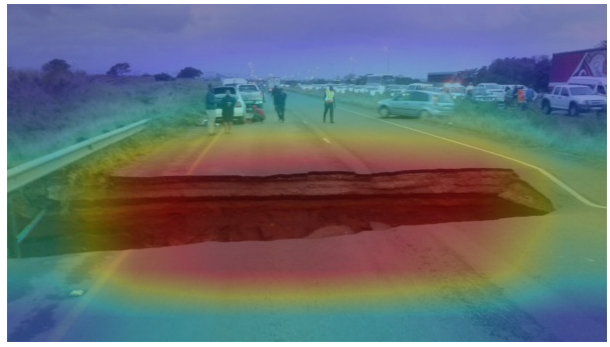}&
\includegraphics[width=\sz,height=\sh]{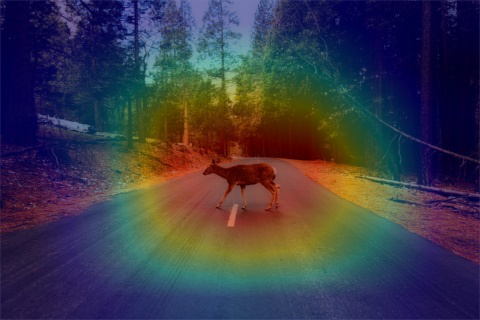}&
\includegraphics[width=\sz,height=\sh]{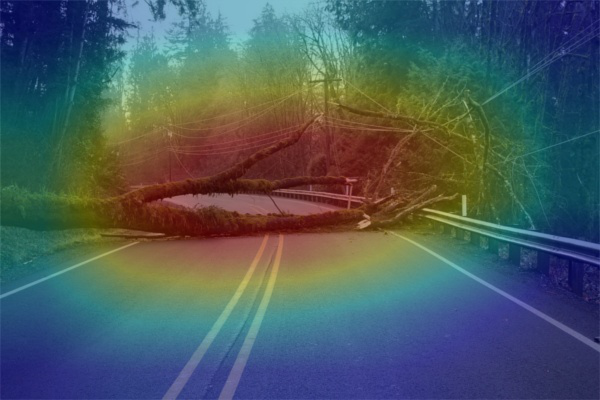}\\
(a) Negative & (b) Collapse & (c) Animal & (d) Treefall \\
 \includegraphics[width=\sz,height=\sh]{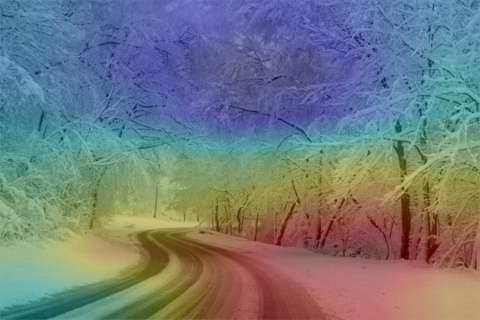}&
\includegraphics[width=\sz,height=\sh]{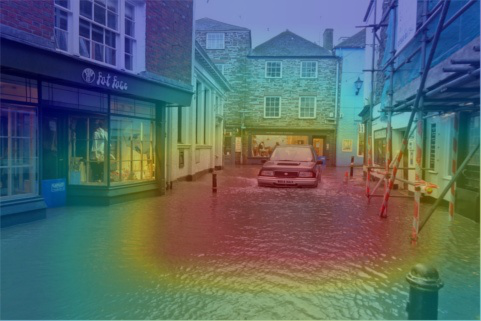}&
\includegraphics[width=\sz,height=\sh]{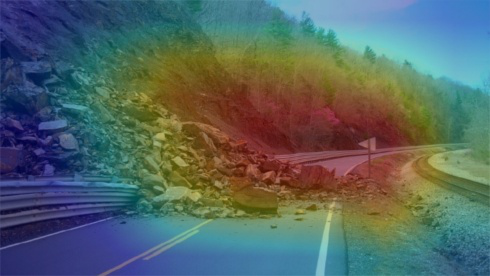}&
 \includegraphics[width=\sz,height=\sh]{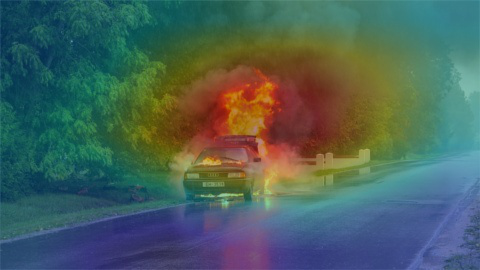}\\
(e) Snow& (f) Flood & (g) Landslide & (h) Fire\\
 \end{tabular}
\caption{Class attention of predicted class overlaid on prototypical images of each class.}
\label{fig:classattention}
\end{figure*}

The confusion matrices show encouraging patterns, and evaluation of the F1-score confirms that the model is not overclassifying images to any particular class. A notably consistent error of the trained model is the confusion between \textit{Snow} or \textit{Flooding} and the negatives class. Both classes have a definition that is hard to delineate, challenging even human classifiers. This uncertainty is reflected in the consistency with which misclassifications occur between the three splits. Notice also how the \textit{Animal on Road} class is hardly ever misclassified during training on the full dataset, but also how it is one of the worst-off classes during validation, despite having a greater amount of training samples compared to other problematic classes such as \textit{Road collapse}. In Figure~\ref{fig:classattention} we inspect the model's visual attention on protoypical incidents by applying Class Activation Mapping, a summation of the total amount of signal at each pixel of the input image. Of the prototypical images, only one incident type is misclassified, namely crashes, classified as a negative scene. The class attention for the other classes is centred on the incident of interest, though the class attention for the class \textit{snow} may drift. Ideally, the class attention for this class should be on the driving surface.

In order to further investigate the model's prediction patterns we apply t-SNE dimension reduction on the last fully-connected layer of the model as in \cite{krizhevsky_imagenet_2012}. t-SNE dimension reduction maps the 512-dimensional vector of the fully-connected layer to just two dimensions for visual interpretation. This is done by clustering points with a high similarity whilst repelling points that are not alike in the original high-dimensional space, then projecting these predictions onto a 2-dimensional plane. Several patterns appear in the t-SNE plots:
\begin{itemize}
\item \textbf{The model can distinguish most classes well}\newline
Most positives classes are clustered together without a fuzzy border towards the grey negatives cluster, with the exception of the \textit{flooding} and the \textit{landslide} classes. This reflect the uncertainty of these classes as per their reported accuracies, while the other classes are far less affected by uncertainty throughout all three splits.
\item \textbf{The \textit{flooding} class has the greatest uncertainty in its classification region}\newline
As indicated by point of interest \textit{d}, the  \textit{flooding} class strongly gravitates towards the Geograph negatives cluster and shares a large indecisive boundary region with it.
\item \textbf{Images within the negatives set are easily distinguishable within the negatives cluster}\newline
Outlined with a red ellipse we find a cluster that predominantly consists of negative Geograph images. We conjecture that this cluster is formed by the comparatively greater amount of countryside images within the Geograph negatives set when compared to both the Berkeley Deep Drive (BDD) and the Cityscapes negatives. 

\begin{figure*}[htbp]
  \centering
    \includegraphics[width=0.8\textwidth]{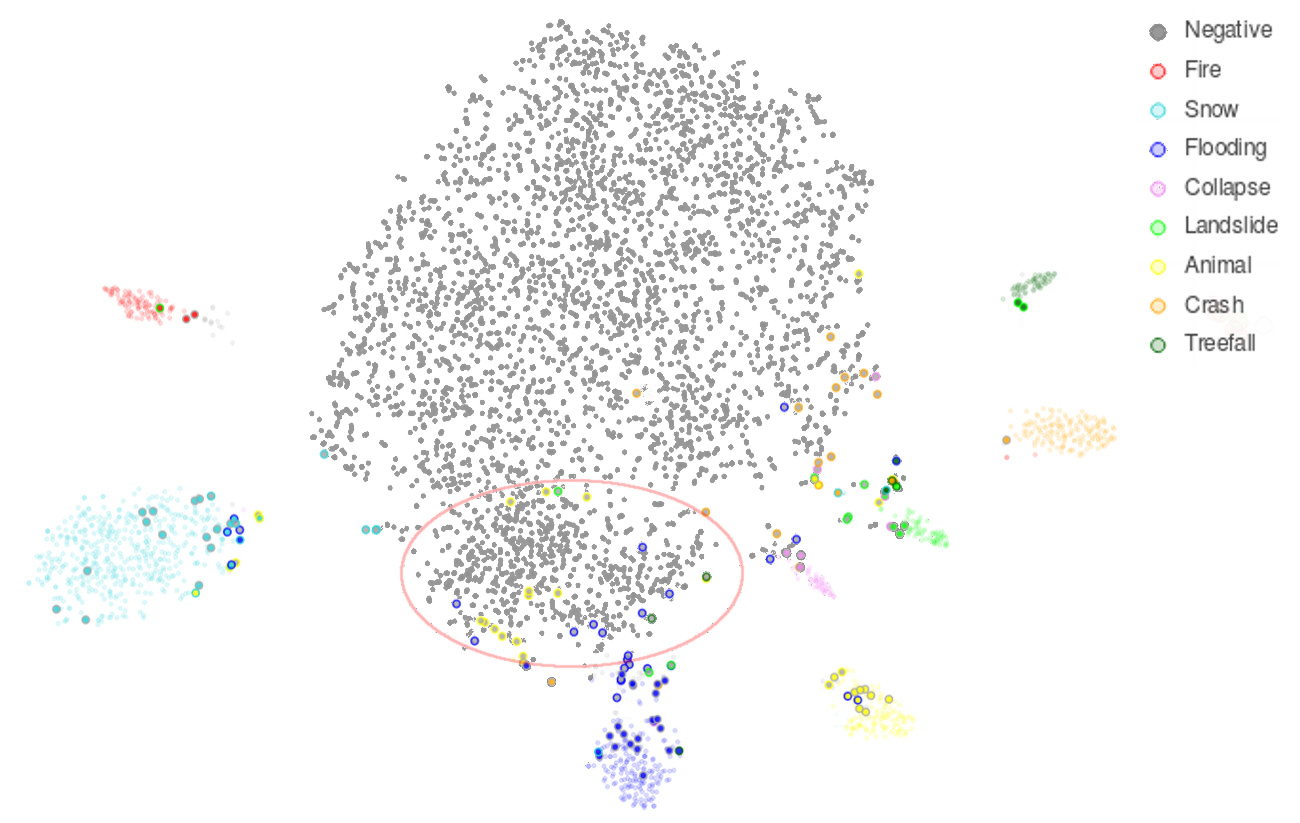}
    \caption{t-SNE dimension reduction of inputs to the fully-connected layer for every image of the complete dataset model, used to make predictions on which class each image belongs to. Plot generated with a perplexity of 50, a learning rate of 500, and a total of 1,000 iterations. The inner circle of each point represents its true class, with the outer circle representing its predicted class. The red ellipse indicates a region consisting almost exclusively of Geograph negatives.}
    \label{fig:tsne_50}
\end{figure*}
\end{itemize}

\subsection{Relevancy/Severity}
\begin{figure}
    \centering
    \begin{tabular}{ccc}
    \includegraphics[width=25mm,height=20mm]{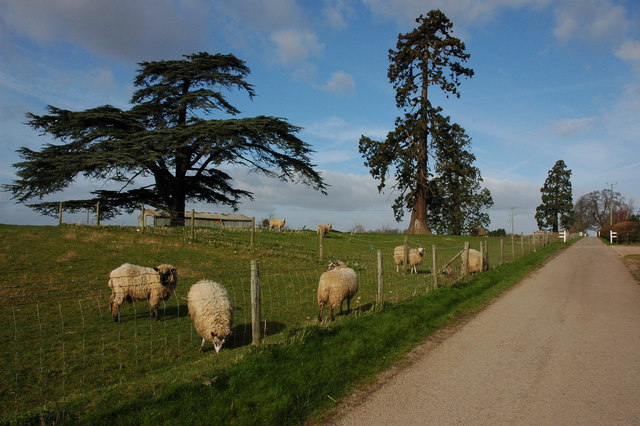}&
    \includegraphics[width=25mm,height=20mm]{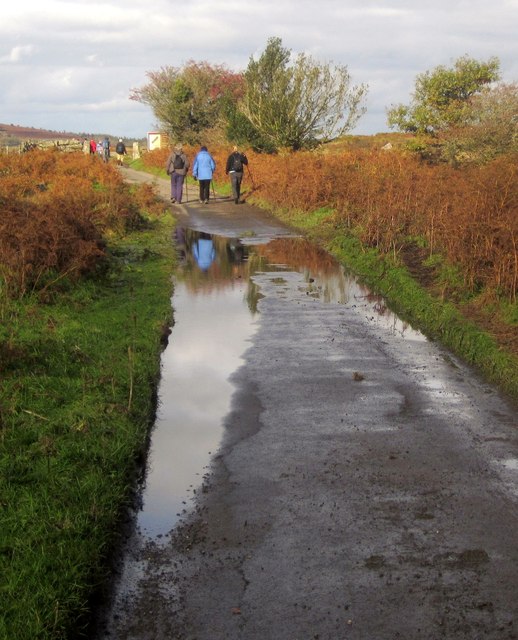}&
    \includegraphics[width=25mm,height=20mm]{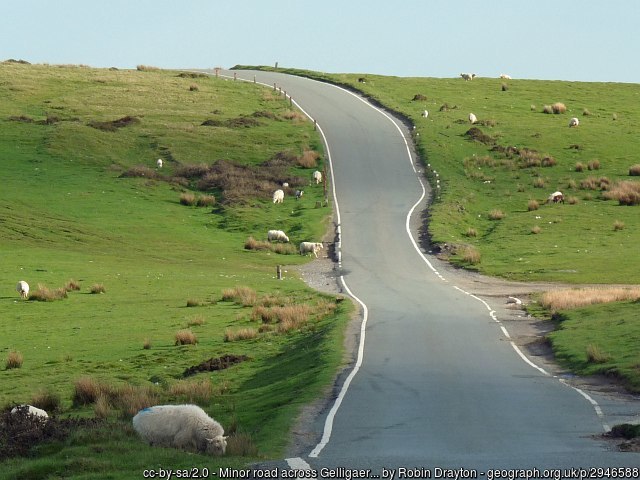}\\
     (a) Animal & (b) Flood & (c) Animal\\
     \end{tabular}
    \caption{Relevancy vs severity. Figure (a) shows an irrelevant scene. Figure (b) shows the difficulty of estimating severity. Figure (c) shows an example where both the relevancy and severity are hard to determine.}
    \label{fig:failures}
\end{figure}

An important focus of future work are the concepts of relevancy and severity. As noted in~\cite{ohn-bar_are_2017}, objects should only be considered incidents if they are spatially relating to the driving situation, thus relevant. Positive class examples gathered during this research all share this feature. %, either explicitly (e.g. \textit{cow is standing on the road}) or implicitly (\textit{sheep standing next to the road}). 
A second characteristic to consider is the severity of the incident, i.e., is an incident significant enough to disrupt road serviceability. In Figure~\ref{fig:failures} we show examples which demonstrate the role of relevancy and severity. We note that both relevancy and severity depend on characteristics of the driving situation such as the type of vehicle used, tire quality, and velocity. As such, the relevancy/severity problem is likely a new complex prediction problem that uses the detection of an incident as input. For the purposes of training, such characteristics can perhaps be averaged for vehicle types to compute type-specific relevance/severity scores for incident examples.

\subsection{Geographical Stratification}
In driving scenarios it is imperative that trained models generalize well across various regions or landscapes. A model needs to detect incidents regardless of the visual properties of the environment. The ability to generalize from training data collected in one region to deployments in unseen geographical regions is therefore an important consideration for a model's fitness to deploy and detection quality. This problem has, to the best of our knowledge, been thus far neglected. Yet, with the advent of self-driving cars with systems trained only on data from limited regions (e.g.,USA), the lack of geographical transfer-ability of training datasets may have serious consequences.

This is especially the case for classifiers for recognizing hazardous situations. If, e.g., an incident detector is unable to recognize a flooding in a desert during a flash-flood, it may result in an autonomous vehicle that drives into a hazardous situation. In this research, we considered the role of geographical correlation between datapoints during training, and we demonstrate the need for future research efforts to consider the adaptability of incident-detection models to unseen geographical domains.

\subsection{Limitations}
\paragraph{Limitations of the Taxonomy of Incidents}
An addition to the process of creating deeper groupings in the taxonomy is to consider the concept of synsets (sets of related synonyms) such as those used in ImageNet \cite{deng_imagenet:_2009}. Princeton's WordNet \cite{miller_wordnet:_1990} may form a good basis for some of the classes in this research which are not combinations of various terms. For instance, \textit{landslide} is listed as a distinct synset, along with its hypernyms \textit{rockslide} and \textit{mudslide}, while \textit{animal on road} is not listed as it is a combination of \textit{animal} and \textit{road}. Instead, the various hypernyms of \textit{animals} may be considered separately, and then combined with their context term (e.g. \textit{road}). For classes which have hypernyms, standardized search terms should be considered as much as possible so that the semantic definition will remain the same throughout future research efforts.

\paragraph{Dataset Limitations}
Images gathered by API harvesting contained many duplicates prior to the selection of images. We filtered many of these duplicates by checking each image with every other image for their exact equivalence without considering resizing and artifacts. This means that resampled and resized images, and images with effect filters applied may be retained in the dataset. While we suspect that only few duplicates remain in the final datasets, it is worth to consider more sophisticated data cleaning approaches in future work, e.g., through feature extraction methods such as perceptual image hashing using feature points, a method able to accurately detect equivalence while accounting for a wide variety of distortions, transformations, and alterations \cite{monga_perceptual_2006}.

Lastly, there is a chance that the full dataset classification may be sensitive to sampling biases as a result of the method for generating the negatives dataset. The negatives set of the full experiment does not contain harvested images which differ from the Geograph negatives and the driving datasets. Thus, the model may have learned to distinguish harvested positives from non-harvested negatives. The geographically stratified model does not suffer from this suspected bias in that the test dataset for this experiment contains the same sources as the training and validation data, and harvested images are only used to enhance the model during the training and validation process. As the accuracy does not differ greatly between both experiments, we do not suspect the likelihood of this bias to be significant. For a decisive test on the influence of biases by source type we suggest that a new experiment is run with a second curated test dataset that also contains harvested samples, so that the degree of bias can be assessed.

\newpage
\section{CONCLUSIONS}\label{sec:Conclusions}
Road networks around the world are under increasing pressure as car ownership rises and road transport intensifies. This increase in road network pressure intensifies the effect that incidents have on the road network. At the same time, vehicles equipped with sensors are becoming increasingly prevalent on the road network as autonomous vehicles are beginning production. To the best of our best knowledge, no existing research has previously been performed on the recognition of incidents as a domain using images as seen in sensor-equipped vehicles. The main motivation for this research was the need to assess the possibility of visual incident detection, as a new class of classification task impacted by constraints to the types of images used, the underspecification of the definition of an incident, and the dependence on geographically stratified large datasets. In this research we have therefore created a taxonomy for unsigned physical incidents, gathered a dataset of images to be used in classification, and confirmed that unsigned physical incidents are learnable by convolutional neural networks with an overall accuracy rate of accuracy rate of 97.15\% and a F1-score of 0.8909. In a second experiment we determined that spatially stratified training and test datasets deteriorate the performance. The overall accuracy of this second experiment was 92.90\% with a F1-score of 0.9169. While this is a small decline showing the model generalizes well, the experiment indicates that the performance in visually very distinct regions would drop rapidly. Future work should therefore focus on a comprehensive collection of geographically distributed training data, to assure consistent performance of such models globally. The dataset is available at \url{https://github.com/Bixbeat/road-incidents-data}

\section*{Acknowledgments}
The authors would like to acknowledge the Geograph UK project for their assistance in using Geograph imagery.

%%%%%%%%%%%%%%%%%%%%%%%%%%%%%% BIBLIOGRAPHY %%%%%%%%%%%%%%%%%%%%%%%%%%%%%%
\bibliographystyle{IEEEtran}
\bibliography{newRefs}

\begin{thebibliography}{10}
\providecommand{\url}[1]{#1}
\csname url@samestyle\endcsname
\providecommand{\newblock}{\relax}
\providecommand{\bibinfo}[2]{#2}
\providecommand{\BIBentrySTDinterwordspacing}{\spaceskip=0pt\relax}
\providecommand{\BIBentryALTinterwordstretchfactor}{4}
\providecommand{\BIBentryALTinterwordspacing}{\spaceskip=\fontdimen2\font plus
\BIBentryALTinterwordstretchfactor\fontdimen3\font minus
  \fontdimen4\font\relax}
\providecommand{\BIBforeignlanguage}[2]{{%
\expandafter\ifx\csname l@#1\endcsname\relax
\typeout{** WARNING: IEEEtran.bst: No hyphenation pattern has been}%
\typeout{** loaded for the language `#1'. Using the pattern for}%
\typeout{** the default language instead.}%
\else
\language=\csname l@#1\endcsname
\fi
#2}}
\providecommand{\BIBdecl}{\relax}
\BIBdecl

\bibitem{highways_england_smart_2016}
\BIBentryALTinterwordspacing
{Highways England}, ``\BIBforeignlanguage{eng}{Smart motorways programme},''
  Jul. 2016. [Online]. Available:
  \url{http://www.highways.gov.uk/smart-motorways-programme/}
\BIBentrySTDinterwordspacing

\bibitem{peluffo_strategic_2015}
\BIBentryALTinterwordspacing
N.~Peluffo, ``Strategic {Road} {Network} {Statistics},'' UK Department of
  Transport, Governmental {Statistics}, Jan. 2015. [Online]. Available:
  \url{https://assets.publishing.service.gov.uk/government/uploads/system/uploads/attachment_data/file/448276/strategic-road-network-statistics.pdf}
\BIBentrySTDinterwordspacing

\bibitem{highways_england_highways_2017}
\BIBentryALTinterwordspacing
{Highways England}, ``Highways {England} {Strategic} {Road} {Network},''
  Highways England, London, Vision document PR129/17, Dec. 2017. [Online].
  Available:
  \url{https://assets.publishing.service.gov.uk/government/uploads/system/uploads/attachment_data/file/666857/Strategic_Road_Network_Initial_Report_Overview.pdf}
\BIBentrySTDinterwordspacing

\bibitem{dargay_vehicle_2007}
\BIBentryALTinterwordspacing
J.~Dargay, D.~Gately, and M.~Sommer, ``Vehicle {Ownership} and {Income}
  {Growth}, {Worldwide}: 1960-2030,'' \emph{The Energy Journal}, vol.~28,
  no.~4, pp. 143--170, 2007. [Online]. Available:
  \url{https://www.jstor.org/stable/41323125}
\BIBentrySTDinterwordspacing

\bibitem{p.b._farradyne_traffic_2000}
{P.B. Farradyne Inc.}, \emph{\BIBforeignlanguage{en}{Traffic {Incident}
  {Management} {Handbook}}}.\hskip 1em plus 0.5em minus 0.4em\relax U.S.
  Department of Transportation, ITS Joint Program Office, 2000,
  google-Books-ID: j3MsAQAAMAAJ.

\bibitem{berdica_introduction_2002}
\BIBentryALTinterwordspacing
K.~Berdica, ``An introduction to road vulnerability: what has been done, is
  done and should be done,'' \emph{Transport Policy}, vol.~9, no.~2, pp.
  117--127, Apr. 2002. [Online]. Available:
  \url{http://www.sciencedirect.com/science/article/pii/S0967070X02000112}
\BIBentrySTDinterwordspacing

\bibitem{zhu_traffic-sign_2016}
Z.~Zhu, D.~Liang, S.~Zhang, X.~Huang, B.~Li, and S.~Hu, ``Traffic-{Sign}
  {Detection} and {Classification} in the {Wild},'' in \emph{2016 {IEEE}
  {Conference} on {Computer} {Vision} and {Pattern} {Recognition} ({CVPR})},
  Jun. 2016, pp. 2110--2118.

\bibitem{zaklouta_real-time_2014}
\BIBentryALTinterwordspacing
F.~Zaklouta and B.~Stanciulescu, ``Real-time traffic sign recognition in three
  stages,'' \emph{Robotics and Autonomous Systems}, vol.~62, no.~1, pp. 16--24,
  Jan. 2014. [Online]. Available:
  \url{http://www.sciencedirect.com/science/article/pii/S0921889012001236}
\BIBentrySTDinterwordspacing

\bibitem{yong_real-time_2015}
H.~Yong and X.~Jianru, ``Real-time traffic cone detection for autonomous
  vehicle,'' in \emph{2015 34th {Chinese} {Control} {Conference} ({CCC})}, Jul.
  2015, pp. 3718--3722.

\bibitem{hillel_recent_2014}
\BIBentryALTinterwordspacing
A.~B. Hillel, R.~Lerner, D.~Levi, and G.~Raz, ``\BIBforeignlanguage{en}{Recent
  progress in road and lane detection: a survey},''
  \emph{\BIBforeignlanguage{en}{Machine Vision and Applications}}, vol.~25,
  no.~3, pp. 727--745, Apr. 2014. [Online]. Available:
  \url{https://link.springer.com/article/10.1007/s00138-011-0404-2}
\BIBentrySTDinterwordspacing

\bibitem{fazekas_locating_2017}
\BIBentryALTinterwordspacing
Z.~Fazekas, G.~Balázs, L.~Gerencsér, and P.~Gáspár, ``Locating roadworks
  sites via detecting change in lateral positions of traffic signs measured
  relative to the ego-car,'' \emph{Transportation Research Procedia}, vol.~27,
  pp. 341--348, Jan. 2017. [Online]. Available:
  \url{http://www.sciencedirect.com/science/article/pii/S2352146517309018}
\BIBentrySTDinterwordspacing

\bibitem{zhou_real-time_2014}
\BIBentryALTinterwordspacing
D.~Zhou, ``\BIBforeignlanguage{en}{Real-time {Animal} {Detection} {System} for
  {Intelligent} {Vehicles}},'' Thesis, Université d'Ottawa / University of
  Ottawa, 2014. [Online]. Available:
  \url{http://ruor.uottawa.ca/handle/10393/31272}
\BIBentrySTDinterwordspacing

\bibitem{saleh_kangaroo_2016}
K.~Saleh, M.~Hossny, and S.~Nahavandi, ``Kangaroo {Vehicle} {Collision}
  {Detection} {Using} {Deep} {Semantic} {Segmentation} {Convolutional} {Neural}
  {Network},'' Nov. 2016, pp. 1--7.

\bibitem{alfarrarjeh_deep_2018}
\BIBentryALTinterwordspacing
A.~Alfarrarjeh, D.~Trivedi, S.~H. Kim, and C.~Shahabi, ``A {Deep} {Learning}
  {Approach} for {Road} {Damage} {Detection} from {Smartphone}
  {Images}.''\hskip 1em plus 0.5em minus 0.4em\relax Seattle, WA: IEEE, Dec.
  2018. [Online]. Available:
  \url{https://www.researchgate.net/publication/330625542_A_Deep_Learning_Approach_for_Road_Damage_Detection_from_Smartphone_Images}
\BIBentrySTDinterwordspacing

\bibitem{chen_lidar-histogram_2017}
L.~Chen, J.~Yang, and H.~Kong, ``Lidar-histogram for fast road and obstacle
  detection,'' in \emph{2017 {IEEE} {International} {Conference} on {Robotics}
  and {Automation} ({ICRA})}, May 2017, pp. 1343--1348.

\bibitem{levi_stixelnet:_2015}
D.~Levi, N.~Garnett, and E.~Fetaya, ``{StixelNet}: {A} {Deep} {Convolutional}
  {Network} for {Obstacle} {Detection} and {Road} {Segmentation},'' Jan. 2015,
  pp. 109.1--109.12.

\bibitem{shao_research_2015}
H.~Shao, Z.~Zhang, and K.~Li, ``Research on water hazard detection based on
  line structured light sensor for long-distance all day,'' in \emph{2015
  {IEEE} {International} {Conference} on {Mechatronics} and {Automation}
  ({ICMA})}, Aug. 2015, pp. 1785--1789.

\bibitem{s._nixon_feature_2012}
M.~S.~Nixon and A.~Aguado, ``Feature {Extraction} \& {Image} {Processing} for
  {Computer} {Vision},'' \emph{Feature Extraction \& Image Processing for
  Computer Vision}, Dec. 2012.

\bibitem{fukushima_neocognitron:_1980}
K.~Fukushima, ``\BIBforeignlanguage{eng}{Neocognitron: a self organizing neural
  network model for a mechanism of pattern recognition unaffected by shift in
  position},'' \emph{\BIBforeignlanguage{eng}{Biological Cybernetics}},
  vol.~36, no.~4, pp. 193--202, 1980.

\bibitem{lecun_gradient-based_1998}
Y.~Lecun, L.~Bottou, Y.~Bengio, and P.~Haffner, ``Gradient-based learning
  applied to document recognition,'' \emph{Proceedings of the IEEE}, vol.~86,
  no.~11, pp. 2278--2324, Nov. 1998.

\bibitem{krizhevsky_imagenet_2012}
\BIBentryALTinterwordspacing
A.~Krizhevsky, I.~Sutskever, and G.~E. Hinton, ``{ImageNet} {Classification}
  with {Deep} {Convolutional} {Neural} {Networks},'' in \emph{Advances in
  {Neural} {Information} {Processing} {Systems} 25}, F.~Pereira, C.~J.~C.
  Burges, L.~Bottou, and K.~Q. Weinberger, Eds.\hskip 1em plus 0.5em minus
  0.4em\relax Curran Associates, Inc., 2012, pp. 1097--1105. [Online].
  Available:
  \url{http://papers.nips.cc/paper/4824-imagenet-classification-with-deep-convolutional-neural-networks.pdf}
\BIBentrySTDinterwordspacing

\bibitem{simonyan_deep_2014}
\BIBentryALTinterwordspacing
K.~Simonyan, A.~Vedaldi, and A.~Zisserman, ``Deep inside convolutional
  networks: visualising image classification models and saliency maps,''
  \emph{International Conference on Learning Representations}, Jan. 2014.
  [Online]. Available:
  \url{https://www.scienceopen.com/document?vid=ebe3c41c-b40c-4ffb-ad87-878dc5a4ad67}
\BIBentrySTDinterwordspacing

\bibitem{he_deep_2016}
\BIBentryALTinterwordspacing
K.~He, X.~Zhang, S.~Ren, and J.~Sun, ``\BIBforeignlanguage{en}{Deep {Residual}
  {Learning} for {Image} {Recognition}},'' in
  \emph{\BIBforeignlanguage{en}{2016 {IEEE} {Conference} on {Computer} {Vision}
  and {Pattern} {Recognition} ({CVPR})}}.\hskip 1em plus 0.5em minus
  0.4em\relax Las Vegas, NV, USA: IEEE, Jun. 2016, pp. 770--778. [Online].
  Available: \url{http://ieeexplore.ieee.org/document/7780459/}
\BIBentrySTDinterwordspacing

\bibitem{hochreiter_vanishing_1998}
S.~Hochreiter, ``The {Vanishing} {Gradient} {Problem} {During} {Learning}
  {Recurrent} {Neural} {Nets} and {Problem} {Solutions},'' \emph{International
  Journal of Uncertainty, Fuzziness and Knowledge-Based Systems}, vol.~6, pp.
  107--116, Apr. 1998.

\bibitem{dixon_roadworks_2017}
\BIBentryALTinterwordspacing
D.~Dixon, ``\BIBforeignlanguage{en}{Roadworks and {Diversion} at
  {Magheralahan}},'' Sep. 2017. [Online]. Available:
  \url{https://www.geograph.ie/photo/5682971}
\BIBentrySTDinterwordspacing

\bibitem{sandy_spring_volunteer_fire_department_ssvfd_2017}
\BIBentryALTinterwordspacing
{Sandy Spring Volunteer Fire Department}, ``\BIBforeignlanguage{en-US}{{SSVFD}
  {Special} {Operations} {Put} to {Work} {During} {Recent} {Flash} {Flood}},''
  May 2017. [Online]. Available:
  \url{https://www.ssvfd.org/ssvfd-special-operations-put-work-recent-flash-flood/}
\BIBentrySTDinterwordspacing

\bibitem{ganter_formal_2012}
B.~Ganter and R.~Wille, \emph{\BIBforeignlanguage{en}{Formal {Concept}
  {Analysis}: {Mathematical} {Foundations}}}.\hskip 1em plus 0.5em minus
  0.4em\relax Springer Science \& Business Media, Dec. 2012, google-Books-ID:
  hNwqBAAAQBAJ.

\bibitem{google_custom_nodate}
\BIBentryALTinterwordspacing
{Google}, ``\BIBforeignlanguage{en}{Custom {Search} {API}}.'' [Online].
  Available: \url{https://developers.google.com/custom-search/}
\BIBentrySTDinterwordspacing

\bibitem{yahoo_flickr_nodate}
\BIBentryALTinterwordspacing
{Yahoo}, ``Flickr {Services} {API} documentation.'' [Online]. Available:
  \url{https://www.flickr.com/services/api/}
\BIBentrySTDinterwordspacing

\bibitem{microsoft_bing_nodate}
\BIBentryALTinterwordspacing
{Microsoft}, ``Bing {Images} {Search} {API} v7 {Reference} {\textbar}
  {Microsoft} {Docs}.'' [Online]. Available:
  \url{https://docs.microsoft.com/en-us/rest/api/cognitiveservices/bing-images-api-v7-reference}
\BIBentrySTDinterwordspacing

\bibitem{seita_bdd100k:_2018}
\BIBentryALTinterwordspacing
D.~Seita, ``{BDD}100k: {A} {Large}-scale {Diverse} {Driving} {Video}
  {Database},'' May 2018. [Online]. Available:
  \url{http://bair.berkeley.edu/blog/2018/05/30/bdd/}
\BIBentrySTDinterwordspacing

\bibitem{cordts_cityscapes_2016}
\BIBentryALTinterwordspacing
M.~Cordts, M.~Omran, S.~Ramos, T.~Rehfeld, M.~Enzweiler, R.~Benenson,
  U.~Franke, S.~Roth, and B.~Schiele, ``\BIBforeignlanguage{en}{The
  {Cityscapes} {Dataset} for {Semantic} {Urban} {Scene} {Understanding}},'' in
  \emph{\BIBforeignlanguage{en}{2016 {IEEE} {Conference} on {Computer} {Vision}
  and {Pattern} {Recognition} ({CVPR})}}.\hskip 1em plus 0.5em minus
  0.4em\relax Las Vegas, NV, USA: IEEE, Jun. 2016, pp. 3213--3223. [Online].
  Available: \url{http://ieeexplore.ieee.org/document/7780719/}
\BIBentrySTDinterwordspacing

\bibitem{noauthor_geograph_2018}
\BIBentryALTinterwordspacing
``\BIBforeignlanguage{en}{Geograph},'' 2018. [Online]. Available:
  \url{http://www.geograph.org.uk/}
\BIBentrySTDinterwordspacing

\bibitem{deng_imagenet:_2009}
J.~Deng, W.~Dong, R.~Socher, L.-j. Li, K.~Li, and L.~Fei-fei, ``Imagenet: {A}
  large-scale hierarchical image database,'' in \emph{In {CVPR}}, 2009.

\bibitem{hinton_neural_2015}
\BIBentryALTinterwordspacing
G.~Hinton, N.~Srivastava, and K.~Swersky, ``Neural {Networks} for {Machine}
  {Learning},'' Toronto, 2015. [Online]. Available:
  \url{https://www.cs.toronto.edu/~tijmen/csc321/slides/lecture_slides_lec6.pdf}
\BIBentrySTDinterwordspacing

\bibitem{google_google_2018}
\BIBentryALTinterwordspacing
{Google}, ``Google {Colaboratory},'' 2018. [Online]. Available:
  \url{https://colab.research.google.com}
\BIBentrySTDinterwordspacing

\bibitem{ohn-bar_are_2017}
\BIBentryALTinterwordspacing
E.~Ohn-Bar and M.~M. Trivedi, ``\BIBforeignlanguage{en}{Are all objects equal?
  {Deep} spatio-temporal importance prediction in driving videos},''
  \emph{\BIBforeignlanguage{en}{Pattern Recognition}}, vol.~64, pp. 425--436,
  Apr. 2017. [Online]. Available:
  \url{http://www.sciencedirect.com/science/article/pii/S0031320316302424}
\BIBentrySTDinterwordspacing

\bibitem{miller_wordnet:_1990}
G.~A. Miller, R.~Beckwith, C.~Fellbaum, D.~Gross, and K.~Miller, ``{WordNet}:
  {An} on-line lexical database,'' \emph{International Journal of
  Lexicography}, vol.~3, pp. 235--244, 1990.

\bibitem{monga_perceptual_2006}
V.~Monga and B.~L. Evans, ``Perceptual {Image} {Hashing} {Via} {Feature}
  {Points}: {Performance} {Evaluation} and {Tradeoffs},'' \emph{IEEE
  Transactions on Image Processing}, vol.~15, no.~11, pp. 3452--3465, Nov.
  2006.

\end{thebibliography}

\end{document}